\documentclass[9pt,conference]{IEEEtran}
\IEEEoverridecommandlockouts
\usepackage{cite}
\usepackage{amsmath,amssymb,amsfonts}
\usepackage{algorithmicx}
\usepackage{graphicx}
\usepackage{textcomp}
\usepackage{xcolor}
\usepackage{pifont}
\usepackage{array}
\usepackage{url}
\usepackage{amsmath} 
\usepackage{bigstrut}
\usepackage[square,sort,comma,numbers,compress]{natbib}
\usepackage{algorithm} 
\usepackage{algpseudocode}
\usepackage{makecell}
\usepackage{bbding}
\usepackage{pifont}
\usepackage{tikz}
\usepackage{titlesec}
\usepackage{pifont}
\usepackage{caption}
\usepackage{flushend}
\usepackage{setspace}

\usepackage{tabularx}
\usepackage{threeparttable}
\usepackage{xspace}
\def\BibTeX{{\rm B\kern-.05em{\sc i\kern-.025em b}\kern-.08em
    T\kern-.1667em\lower.7ex\hbox{E}\kern-.125emX}}

\def\framework{Muffin}
\usepackage{multirow}

\begin{document}

\title{
\framework: A Framework Toward Multi-Dimension AI Fairness by Uniting Off-the-Shelf Models\\
}


\author{
\IEEEauthorblockN{
\Large
Yi Sheng$^{1}$ \quad
Junhuan Yang$^{1}$ \quad
Lei Yang$^{1}$ \quad
Yiyu Shi$^{2}$ \quad
Jingtong Hu$^{3}$ \quad
Weiwen Jiang$^{1}$ \quad
}
\IEEEauthorblockA{
\Large
$^{1}$ George Mason University \quad
$^{2}$ University of Notre Dame \quad
$^{3}$ University of Pittsburgh
}}

\maketitle

\begin{abstract}

Model fairness (a.k.a., bias) has become one of the most critical problems in a wide range of AI applications. An unfair model in autonomous driving may cause a traffic accident if corner cases (e.g., extreme weather) cannot be fairly regarded; or it will incur healthcare disparities if the AI model misdiagnoses a certain group of people (e.g., brown and black skin).
In recent years, there are emerging research works on addressing unfairness, and they mainly focus on a single unfair attribute, like skin tone; however, real-world data commonly have multiple attributes, among which unfairness can exist in more than one attribute, called ``multi-dimensional fairness''.
In this paper, we first reveal a strong correlation between the different unfair attributes, i.e., optimizing fairness on one attribute will lead to the collapse of others.
Then, we propose a novel \underline{Mu}lti-Dimension \underline{F}a\underline{i}r\underline{n}ess framework, namely \textit{Muffin}, which includes an automatic tool to unite off-the-shelf models to improve the fairness on multiple attributes simultaneously.
Case studies on dermatology datasets with two unfair attributes show that 
the existing approach can achieve 21.05\% fairness improvement on the first attribute while it makes the second attribute unfair by 1.85\%.
On the other hand, the proposed \textit{Muffin} can unite multiple models to achieve simultaneously 26.32\% and 20.37\% fairness improvement on both attributes; meanwhile, it obtains 5.58\% accuracy gain.






\end{abstract}

\section{Introduction}

With the continuous progress of AI democratization, we have  witnessed the employment of machine learning (ML) in various applications \cite{yang2023device,sheng2022larger,peng2022towards,liao2021shadow,yang2023pruning,yan2023viability,peng2023autorep,huang2021path}, 
such as autonomous driving \cite{dogan2011autonomous}, language translation \cite{corston2001machine}, dermatology assistant \cite{rambhajani2015survey}, and vital signs monitoring \cite{chen2005mobile}. 
Along with such progress, the fairness issue in ML models has emerged and become critical in applications. 
An unfair ML model performs well in most cases, but its performance may crash when corner cases occur.
This may lead to accidents in autonomous driving \cite{accident}, linguistic discrimination in language translation \cite{gonzalez2013translating}, and even threatening life by misdiagnosis in healthcare \cite{suite2007beyond}.
Thus, it is high time to take fairness into consideration in designing the ML system.


In recent years, a notable number of research works aim to address the AI fairness issue from an algorithm's perspective \cite{holstein2019improving}; however, there are very few works taking this problem from a systematic view that can provide an automatic tool to generate a fair ML system.
In the existing works, there are commonly two directions: 
(1) Revising training algorithm by using adversarial training \cite{xu2021robust}, domain discriminative training \cite{wang2020towards}, or training with a fair loss function \cite{liu2019fair}; and (2) Improving data balance by generating an unprivileged group of data using GAN \cite{sattigeri2019fairness}. 
These methods have achieved a great improvement in the fairness of a single unfair attribute; however, almost all of these works ignore the important fact that one dataset will have several fairness-related attributes, called ``multi-dimensional fairness''.
For example in the dermatology dataset, the unfairness may not only occur at the skin tone attribute, but also at age, gender, or disease site.
It seems straightforward to extend the existing approach; however, by examining the state-of-the-art solutions that optimize fairness from the algorithm's perspective \cite{holstein2019improving}, we observe that there exists a correlation between unfair attributes, that is, improving fairness on one attribute will decrease that on the other one.

The correlation among unfair attributes creates a challenge of achieving multi-dimensional fairness by optimizing a single model.
In response to this, we take an alternative way to leverage and unite multiple existing models to improve fairness.
This is inspired by our key observation that two models with similar accuracy on an unprivileged group may have a disagreement on specified data, that is, for one data, only one model makes the correct classification.
As a result, if we can correct the results by smartly selecting the model to be used when a disagreement happens, then we can improve the fairness of an unprivileged group.
Although promising, it is still unclear which models should be paired, how can we correct classification. when a disagreement happens, and how to improve the fairness of multiple attributes.

In this paper, we propose a novel framework, namely ``Muffin'', to address all of the above issues in a systematic way.
Given a dataset with multiple unfair attributes and a set of off-the-shelf models, the proposed framework can automatically select multiple models to maximize the fairness of all attributes simultaneously.
To achieve this goal, we first develop a model-fusing structure, where a multi-layer perception (MLP) will take the output of the selected models as input.
Then, we will create a proxy dataset by assigning weights to data with multiple attributes in unprivileged groups.
On top of this, we employ the reinforcement learning approach to identify the model-fusing structure (i.e., select the models to be involved and search for the best MLP architecture).
The identified architecture will be trained on the proxy dataset to achieve multi-dimension fairness.

The main contributions of this paper are as follows.
 
\begin{itemize}
  \item \textbf{Problem.} To the best of our knowledge, this is the first work to identify the multi-dimensional fairness problems in the ML domain. We reveal the correlation among different unfair attributes, which inhibits us from directly applying existing single-dimensional fairness techniques. 
  \item \textbf{Framework.} Given a dataset with multiple unfair attributes and a set of ML models, a novel framework, namely Muffin, is proposed to unite off-the-shelf models to improve multiple unfair attributes simultaneously. 
  \item \textbf{Automated Tool.} A reinforcement learning-based algorithm coupled with a fairness-aware proxy dataset is developed to enable the optimizations to be performed automatically.
\end{itemize}

Experimental results on the two dermatology datasets evaluate the
performance of Muffin. 
First, with small models like ShuffleNet\_V2\_X1\_0, and MobileNet\_V3\_Small, Muffin can improve the overall accuracy to 80.55\% and 81.77\% from 77.21\% and 76.19\%, respectively.
For ShuffleNet\_V2\_X1\_0, Muffin can achieve 19.44\% and 2.22\% improvements on two unfair attributes; while the figures are 26.32\% and 20.37\% for MobileNet\_V3\_Small.
With large models, the overall accuracy can be maintained by Muffin; meanwhile, it can consistently improve unfair attributes by 7.69\% and 9.30\% for ResNet-18.



In the rest of the paper, Section 2 reviews the related work
and provides motivation; Section 3 defines the problem and
presents Muffin. Experimental results are reported
in Section 4 and concluding remarks are given in Section 5.

\begin{figure}[t]
  \centering
  \includegraphics[width=3.4in]{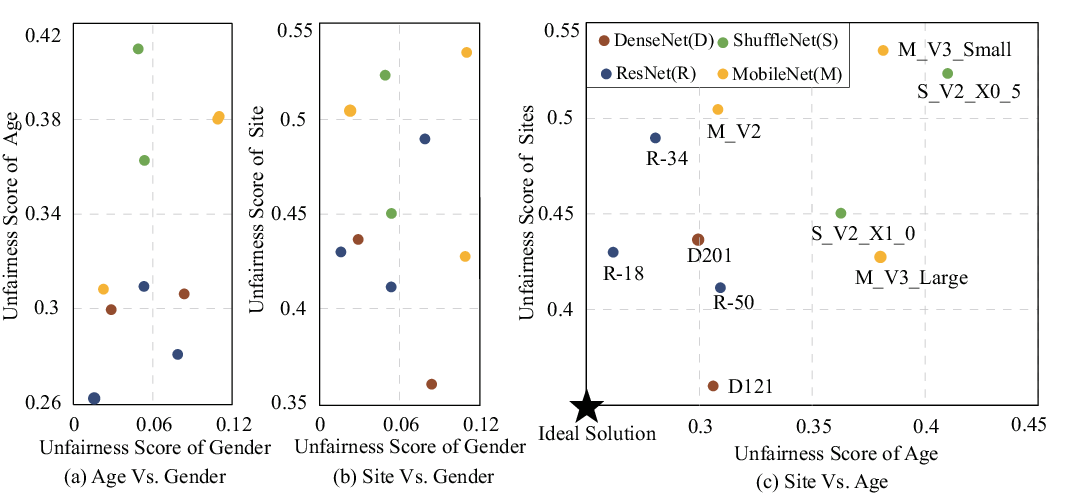}
  \caption{Fairness of existing neural architectures on different attributes: (a-b) unfairness score on gender has small variance; (c) both site and age attributes have high unfairness score.}
  \label{fig:mot1}
\end{figure}

\section{Related Work and Motivation}
\label{sec:pre}

This section provides several key observations on AI fairness and discusses the existing works, which gives us the motivation to research the fairness issue on multiple attributes.

\textbf{Observation 1: Unfairness exists on multiple attributes.}

On the dermatology dataset ISIC2019 \cite{ISIC2019}, there are three attributes (age, gender, and disease site). 
We examine the fairness of these attributes by using a set of commonly used neural architectures, including MobileNet \cite{sandler2018mobilenetv2}, ResNet \cite{targ2016resnet}, DenseNet \cite{huang2017densely}, and ShuffleNet \cite{ma2018shufflenet}.
After training these architectures on ISIC2019 dataset, Figure \ref{fig:mot1} shows the unfairness score of the trained model on different attributes.
In Figure \ref{fig:mot1}(a)-(b), we compare the fairness of different models between the attributes of age/site and gender.
We can see that the maximum unfairness score of all models on the gender attribute is less than 0.12, which reflects the low accuracy gap between males and females (i.e., merely 3\%).
On the other hand, in Figure \ref{fig:mot1}(c), we observe that both attributes age and site have high unfairness scores (i.e., over 0.4), which are led by 45.04\% and 36.27\% accuracy gaps between different groups in site and age attributes, respectively.
More importantly, the unfairness scores of these two attributes on different models are not positively correlated. For example, 
regarding the unfairness score of site, DenseNet121 (denoted D121) has the best performance; however, when it comes to age, ResNet-18 (denoted R18) has the lowest unfairness score. 
Results demonstrated that there is no neural architecture can take over both age and site.

\textbf{{Related work: Fairness on multiple attributes has not been well studied in existing research works.}}


Most AI fairness research efforts target a single attribute.
Regarding to the technique, there are commonly two directions to improve fairness: (1) balancing data, and (2) improving training.
First, data unbalancing can easily lead to unfairness; as a result, the most straightforward way is to balance data by applying data augmentation (such as flip, rotation, and scale) on the unprivileged groups \cite{sharma2020data,vannur2021data,iosifidis2018dealing}.
Second, since the unfairness is measured by the accuracy of different groups of data and the accuracy is directly affected by training, the methods to improve training for high fairness were proposed \cite{liu2019fair,du2021fairness,madras2018predict}.
The intuition is that fairness can be improved if we can add a regularization term in the loss function with the consideration of bias.
Besides data and training, authors in \cite{sheng2022larger} also point out that neural network architecture can also affect model fairness.

All the above works aimed at optimizing fairness on a single attribute, when it comes to multiple attributes, it is unclear whether we can directly apply these approaches to optimize the model with several unfair attributes simultaneously.

\begin{figure}[t]
  \centering
  \includegraphics[width=3.4in]{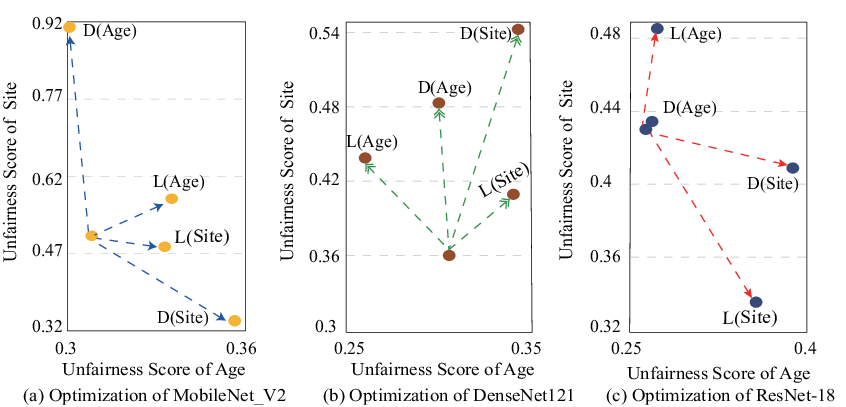}
  \caption{No existing methods can improve two unfair attributes simultaneously. }
  \label{fig:mot2}
\end{figure}




\textbf{{Observation 2: Fairness attributes are entangled and models encounter bottlenecks in improving fairness.}}

To answer the above question, we conduct a set of experiments using both balancing data ($D$) and improving training by a fair loss function ($L$) on the ISIC2019 dataset.
For the neural network, we select three networks, including R18 and D121 on the Pareto frontier in Figure \ref{fig:mot1}(c), and M\_V2.
Results are reported in Figure \ref{fig:mot2}.


From Figure \ref{fig:mot2}(a), we observe that 
two fairness-related attributes (i.e., age and site) are entangled with each other. 
Specifically, when we optimize age, the unfairness score of the site will increase (see $D(Age)$). Correspondingly, if we apply the optimization method to the site, the model on the age attribute will become unfairer (see $L(Site)$ and $D(Site)$). It's like a seesaw game, we can only press down one end but not both ends at the same time. 

From Figure \ref{fig:mot2}(b)-(c), we further observe that it is difficult to improve the model with a low fairness score.
For example, DenseNet121 in Figure \ref{fig:mot2}(b) has a low unfairness score on site; we cannot further reduce this score, even if we use $D$ and $L$ on the site attribute.
Similar behavior occurs on the age attribute of the ResNet-18 model, as shown in Figure \ref{fig:mot2}(c).
In consequence, 
neural network models will encounter bottlenecks when doing the single attribute optimization.
The above results demonstrate the limits of the single attribute optimization and motivate us to find a solution to solve multi-dimensional fairness.

\begin{figure}[t]
  \centering
  \includegraphics[width=3.2in]{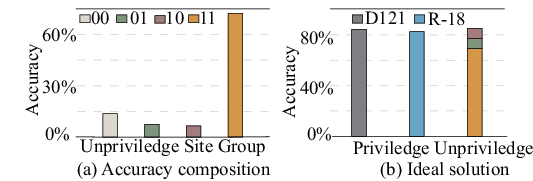}
  \caption{Probability for unprivileged group with resnet18 and optimized denseNet121 for attribute site. Note: 00 indicates both models are doing the wrong classification; 01 means ResNet-18 is correct and DenseNet121 is incorrect; 10 is the opposite; 11 means both models have correct classification.}
    \vspace{-8pt}
  \label{fig:mot3_b}
\end{figure}

{\textbf{Observation 3: Models are complementary on fairness.}}
From the previous example, it seems that optimizing fairness on multiple attributes is unattainable.
In this paper, we consider such a multi-dimensional fairness problem from another angle: Instead of optimizing one model, can we leverage multiple models to improve fairness?
We breakdown the classification results of R18 and D121 in 
Figure \ref{fig:mot3_b}.
As shown in Figure \ref{fig:mot3_b}(a),  two middle bars indicate the probability of one model is correct and another one is incorrect, of which the sum probability reaches 15.93\%. 
This inspires us to leverage the correct result of multiple models (called unite multiple models), and then we can improve the accuracy of the unprivileged group.
Figure \ref{fig:mot3_b}(b) further shows that if we can unite two models for the unprivileged group, its accuracy can be even higher than the accuracy of privileged groups on both models.

{\textbf{Motivation: Unite off-the-shelf models to improve fairness on multiple attributes at the same time.}}

With the insight that multiple models can improve fairness, given a set of off-the-shelf models (called model pool), we propose an automatic tool named ``Muffin'' to design the system with the objective of maximum fairness on multiple unfair attributes.
Specifically, the proposed structure contains two components: (1) the ``muffin body'' will be made by selecting models from the model pool; (2) the ``muffin head'' is based on a multi-layer perceptron to process the results of multiple models, which are expected make a correct classification when multiple models have a disagreement. 
We will formally formulate the problem and provide the detailed design of the Muffin framework in the next section.

\section{Muffin Framework}
In this paper, we study the multi-dimensional fairness for classification tasks in computer vision. 
This section will formally define the problem of ``Optimization of multi-dimensional fair classification'' and gives out framework of Muffin.

\noindent\textbf{3.1 Problem Definition}

\textbf{Classification.} Given a dataset $D$, we define $C=\{c_1,c_2,\cdots,c_M\}$ as a set of $M$ classes, where each data $d_i\in D$ belongs to a class $c_j\in C$. That is, there exists a mapping function $f$: $f(d_i)=c_j$. A network $N$ is building it from $D$ to $C$.
With training data, $N$ will learn a function $f_N^{\prime}$ to approximate $f$. If $f(d_i)=f_N^{\prime}(d_i)$, it is a correct classification on data $d_i$. The accuracy $A(f_N^{\prime},D)$ describes the ratio of data in $D$ getting the correct classification by model $N$.

\textbf{Attribute.} For a dataset $D$, we define $a_k$ as a single sensitive attribute.  For each $a_k$, the whole dataset has $G$ groups $D=\{D_1,D_2,\cdots,D_G\}$. 
Each data $d_i\in D$ has the sensitive attribute set $A=\{a_1,a_2,\cdots,a_K\}$. That is the dataset $D$ can be divided according to different attributes $a_k$.  


\textbf{Multi-Dimensional Fairness.} Fairness reflects the model's performance of diverse groups $D_g$ for the single attribute $a_k$. For a model $N$, the accuracy for one group is $A(f_N^{\prime},D_g)$. Based on the accuracy, we define the unfairness score $U$ of a model $N$ for attribute $a_k$ on the dataset $D$ based on L1-norm,which is $U(f_N^\prime,D)_{a_k}=\sum_{\forall g\in G}\{|A(f_N^{\prime},D_{g})_{a_k}-A(f_N^\prime,D)_{a_k}|\}$.
When it comes to the multi-dimensional fairness score $U$, one can extend the fairness from one single attribute to multiple attributes by formulating the objective as:
\vspace{-1em}
\begin{equation}
\small
   U=\sum_{k=1}^{K} U(f_N^\prime,D)_{a_k}
\end{equation}

\vspace{-6pt}
\textbf{Problem Formulation.} Based on the above definitions, we can formally define the problem as follows: Given a dataset $D$ which has an attribute set $A$
with $M$ classes and off-the-shelf models, our objective is to automatically optimize several unfair attributes at one time, it means we need to minimize the unfairness score $U(f_N^\prime, D)_{a_k}$ of each unfair attribute and the multi-dimensional unfairness score $U$, meanwhile overall accuracy $A(f_N^\prime, D)$ meet the requirement.

\begin{figure}[t]
  \centering
  \includegraphics[width=3.4in]{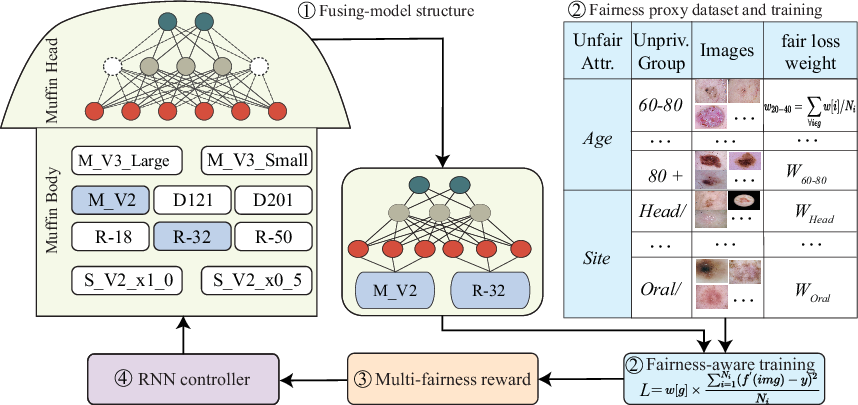}
  \caption{Overview of muffin framework.}
  
  \label{fig:framework}
    \vspace{-5pt}
\end{figure}

\noindent\textbf{3.2 Framework Overview}

Figure \ref{fig:framework} illustrates the overview of our proposed Muffin framework. 
It is composed of four components: \ding{192} model-fusing structure based on a model pool and a backbone structure of MLP, \ding{193} a fairness proxy dataset and training, \ding{194} a multi-fairness reward, and \ding{195} a recurrent neural network (RNN) based controller. 

These components will be iteratively executed for optimization.
A model-fusing structure will be identified from the search space defined in component \ding{192}. Then,
it will be sent to component \ding{193} for training. 
To enable the optimization of fairness on multiple unfair attributes, we will build a fairness proxy dataset, which will be utilized in training.
Next, the trained model will be used to generate a multi-fairness reward in component \ding{194}, and \ding{195} controller will be updated based on the reward in each iteration.
In the following text, we will introduce these components in detail.

{\large\ding{192}} \textbf{Model-Fusing Structure}: 
It contains two parts: (1) ``Muffin body" includes a set of off-the-shelf models, and these models can be selected by the controller to be utilized in the final structure; (2) in ``Muffin head'', an 
MLP backbone specifies a set of hyperparameters, including the number of layers, the number of parameters in each layer, and different activation functions. 
The generated model-fusing structure will be composed of the selected models in the muffin body and the identified MLP from the muffin head, where the MLP is expected to be trained such that it can correct select results when there has a disagreement on multiple selected models.


\vspace{1pt}
{\large\ding{193}} \textbf{Fairness Proxy Dataset and Training}: 
In order to train the model-fusing structure, there are two fundamental questions: which parameters will be tuned, and what dataset will be used?
To maximize the search efficiency, we will freeze the parameters in the pre-trained off-the-shelf models in the model-fusing structure and train parameters in MLP only.
For the second question, we will only involve unprivileged group data for training.
This is based on the \textbf{Observation 3} in section \ref{sec:pre}, where the proposed technique is not going to change the output if all models reached consensus.
In our experiments, we further observe that models in privileged group data have a low rate of disagreement.
Therefore, we will not involve 
the privileged group data for model-fusing structure training. 

The next question is how to consider multiple unfair attributes in training.
As pointed out by \cite{yan2022forml}, one of the major challenges for fairness is how to improve the impact of biased data which are difficult to recognize in training.
Based on this concept, we carry out a one-time pre-processing to consider multiple unfair attributes.
More specifically, we will assign a weight for each group, and it will be used to calculate the loss in training.
In the following algorithm, we give the pseudo code to generate the weights.

\begin{algorithm}
\footnotesize
	\caption{Calculation of weight}    
	 \label{weight}      
	\begin{algorithmic}[0]
	\Require  The output training data for unprivileged groups by paired models  
    \Ensure  Weight of every unprivileged  group for each unfair attribute 
    \State $a_k$ unfair attribute, $A$ the set of the unfair attributes, $g$ every unprivileged group, $G$ unprivileged group set, $N_i$ the  number of images in the unprivileged group 

    \For {$a_k\in A$, $g\in G_{a_k}$, img$\in g$:}      
    
        w[img]+=1; // w is the weight for every image
   \EndFor
    
    \For {$a_k\in A$, $g\in G\_{a_k}$, img$\in g$:}      

        w[g]=$\sum_{\forall i\in g}$w[i]/$N_i$;// w is the weight for every unprivileged group
      
   \EndFor
   
   \end{algorithmic}
   \label{al1}
\end{algorithm} 
In Algorithm \ref{al1}, 
we consider the impact of the same data on the different unprivileged groups, and we associate the data with a higher weight if it appears in the groups under multiple unfair attributes.
In this way, the multiple unfair attributes can be considered in a holistic optimization.
Based on the obtained weights, we formulate the loss function in our fairness-aware training as follows:
\begin{equation}
L= w[g]\times \frac{\sum_{i=1}^{N   }(f^{'}(x)-y )^{2}  }{N }  
\end{equation}

\vspace{1pt}
{\large\ding{194}} \textbf{Multi-Fairness Reward}: After the training is finished, our next target is to generate a reward to assess the identified model.
To this end, we will perform inference on the original dataset using the trained model-fusing structure to 
generate overall accuracy $A(f_N^\prime,D)$ and unfairness score $U(f_N^\prime,D)_{a_k}$ for each unfair attribute.
Then, the reward $R$ is generated based on these metrics, which can be formulated as below.

\vspace{-1.5em}
\begin{equation}\label{equ:reward}
\small
Reward=\sum_{k=1}^{K} \frac{A(f^{\prime},D)}{U(f_N^\prime,D)_{a_k}} 
\end{equation}
where $K$ is the number of unfair attributes, and $a_k$ is a specific attribute.
A larger reward indicates the model has higher accuracy and lower unfairness on average.

{\large\ding{195}} \textbf{RNN Controller}:
The controller will iteratively generate a model-fusing structure by determining the hyperparameters using a recurrent neural network (RNN).
In each step of RNN, a fully connected (FC) layer with parameters will generate one hyperparameter.
And the parameters in RNN and FC layers will be updated by using the reward.
More specifically, we apply the Monte Carlo policy gradient algorithm \cite{williams1992simple}:

\begin{equation}
\small
    \nabla J(\theta) = \frac{1}{m}\sum_{k=1}^{m}\sum_{t=1}^{T}\gamma^{T-t}\nabla_{\theta}\log \pi_{\theta} (a_{t}|a_{(t-1):1})(R_{k}-b)
\end{equation}

where $m$, $T$, $\gamma$, and $b$ indicate the batch size, the number of steps in each episode, an exponential factor, and an average exponential moving of rewards, respectively.
In every episode, the reward is discounted by
$\gamma$ and $b$.

\begin{table*}[]
\centering
\footnotesize
\renewcommand\arraystretch{1.3}
\tabcolsep 1.3pt
  \caption{Comparison of Muffin with existing fairness techniques for different architectures}
\begin{tabular}{|c|ccc|ccc|ccc|cccccccc|}
\hline
                                        & \multicolumn{3}{c|}{Villina(vil.)}                                                                                    & \multicolumn{3}{c|}{Method $D$\cite{weiss2007cost}}                                                                                        & \multicolumn{3}{c|}{Method $L$ \cite{jozani2006estimation}}                                                                                        & \multicolumn{8}{c|}{Muffin}                                                                                                                                                                                                                                                                                                                                                                                                          \\ \cline{2-18} 
\multirow{-2}{*}{Model}                 & \multicolumn{1}{c|}{Age}                    & \multicolumn{1}{c|}{Site}                   & Acc.                  & \multicolumn{1}{c|}{Age}                         & \multicolumn{1}{c|}{Site}                        & Acc. & \multicolumn{1}{c|}{Age}                         & \multicolumn{1}{c|}{Site}                        & Acc. & \multicolumn{1}{c|}{MLP}                                & \multicolumn{1}{c|}{Paired Models}                 & \multicolumn{1}{c|}{Age}                                           & \multicolumn{1}{c|}{vs.Vil}                    & \multicolumn{1}{c|}{Site}                                          & \multicolumn{1}{c|}{vs.Vil.}                   & \multicolumn{1}{c|}{Acc.}                  & Acc.Imp.                 \\ \hline
                                        & \multicolumn{1}{c|}{}                       & \multicolumn{1}{c|}{}                       &                           & \multicolumn{1}{c|}{{\color[HTML]{32CB00} 0.28}} & \multicolumn{1}{c|}{{\color[HTML]{FE0000} 0.49}} & 78.32\%  & \multicolumn{1}{c|}{{\color[HTML]{32CB00} 0.31}} & \multicolumn{1}{c|}{{\color[HTML]{FE0000} 0.57}} & 74.24\%  & \multicolumn{1}{c|}{}                                   & \multicolumn{1}{c|}{}                              & \multicolumn{1}{c|}{{\color[HTML]{32CB00} }}                       & \multicolumn{1}{c|}{}                          & \multicolumn{1}{c|}{{\color[HTML]{32CB00} }}                       & \multicolumn{1}{c|}{}                          & \multicolumn{1}{c|}{}                          &                          \\ \cline{5-10}
\multirow{-2}{*}{ShuffleNet\_V2\_X1\_0} & \multicolumn{1}{c|}{\multirow{-2}{*}{0.36}} & \multicolumn{1}{c|}{\multirow{-2}{*}{0.45}} & \multirow{-2}{*}{77.21\%} & \multicolumn{1}{c|}{{\color[HTML]{32CB00} 0.30}} & \multicolumn{1}{c|}{{\color[HTML]{FE0000} 0.68}} & 77.49\%  & \multicolumn{1}{c|}{{\color[HTML]{FE0000} 0.46}} & \multicolumn{1}{c|}{{\color[HTML]{FE0000} 0.66}} & 76.70\%  & \multicolumn{1}{c|}{\multirow{-2}{*}{{[}16,18,12,8{]}}} & \multicolumn{1}{c|}{\multirow{-2}{*}{DenseNet121}} & \multicolumn{1}{c|}{\multirow{-2}{*}{{\color[HTML]{32CB00} 0.29}}} & \multicolumn{1}{c|}{\multirow{-2}{*}{19.44\%}} & \multicolumn{1}{c|}{\multirow{-2}{*}{{\color[HTML]{32CB00} 0.44}}} & \multicolumn{1}{c|}{\multirow{-2}{*}{2.22\%}}  & \multicolumn{1}{c|}{\multirow{-2}{*}{80.55\%}} & \multirow{-2}{*}{3.34\%} \\ \hline
                                        & \multicolumn{1}{c|}{}                       & \multicolumn{1}{c|}{}                       &                           & \multicolumn{1}{c|}{{\color[HTML]{32CB00} 0.30}} & \multicolumn{1}{c|}{{\color[HTML]{FE0000} 0.55}} & 77.90\%  & \multicolumn{1}{c|}{{\color[HTML]{34FF34} 0.33}} & \multicolumn{1}{c|}{{\color[HTML]{FE0000} 0.69}} & 76.09\%  & \multicolumn{1}{c|}{}                                   & \multicolumn{1}{c|}{}                              & \multicolumn{1}{c|}{{\color[HTML]{32CB00} }}                       & \multicolumn{1}{c|}{}                          & \multicolumn{1}{c|}{{\color[HTML]{32CB00} }}                       & \multicolumn{1}{c|}{}                          & \multicolumn{1}{c|}{}                          &                          \\ \cline{5-10}
\multirow{-2}{*}{MobileNet\_V3\_Small}  & \multicolumn{1}{c|}{\multirow{-2}{*}{0.38}} & \multicolumn{1}{c|}{\multirow{-2}{*}{0.54}} & \multirow{-2}{*}{76.19\%} & \multicolumn{1}{c|}{{\color[HTML]{FE0000} 0.40}} & \multicolumn{1}{c|}{{\color[HTML]{FE0000} 0.63}} & 76.46\%  & \multicolumn{1}{c|}{{\color[HTML]{FE0000} 0.44}} & \multicolumn{1}{c|}{{\color[HTML]{FE0000} 0.64}} & 73.09\%  & \multicolumn{1}{c|}{\multirow{-2}{*}{{[}16,10,10,8{]}}} & \multicolumn{1}{c|}{\multirow{-2}{*}{ResNet-34}}   & \multicolumn{1}{c|}{\multirow{-2}{*}{{\color[HTML]{32CB00} 0.28}}} & \multicolumn{1}{c|}{\multirow{-2}{*}{26.32\%}} & \multicolumn{1}{c|}{\multirow{-2}{*}{{\color[HTML]{32CB00} 0.43}}} & \multicolumn{1}{c|}{\multirow{-2}{*}{20.37\%}} & \multicolumn{1}{c|}{\multirow{-2}{*}{81.77\%}} & \multirow{-2}{*}{5.58\%} \\ \hline
                                        & \multicolumn{1}{c|}{}                       & \multicolumn{1}{c|}{}                       &                           & \multicolumn{1}{c|}{{\color[HTML]{32CB00} 0.30}} & \multicolumn{1}{c|}{{\color[HTML]{FE0000} 0.48}} & 82.44\%  & \multicolumn{1}{c|}{{\color[HTML]{34FF34} 0.26}} & \multicolumn{1}{c|}{{\color[HTML]{FE0000} 0.44}} & 81.65\%  & \multicolumn{1}{c|}{}                                   & \multicolumn{1}{c|}{}                              & \multicolumn{1}{c|}{{\color[HTML]{32CB00} }}                       & \multicolumn{1}{c|}{}                          & \multicolumn{1}{c|}{{\color[HTML]{32CB00} }}                       & \multicolumn{1}{c|}{}                          & \multicolumn{1}{c|}{}                          &                          \\ \cline{5-10}
\multirow{-2}{*}{DenseNet121}           & \multicolumn{1}{c|}{\multirow{-2}{*}{0.31}} & \multicolumn{1}{c|}{\multirow{-2}{*}{0.36}} & \multirow{-2}{*}{81.83\%} & \multicolumn{1}{c|}{{\color[HTML]{FE0000} 0.34}} & \multicolumn{1}{c|}{{\color[HTML]{FE0000} 0.54}} & 83.31\%  & \multicolumn{1}{c|}{{\color[HTML]{FE0000} 0.34}} & \multicolumn{1}{c|}{{\color[HTML]{FE0000} 0.41}} & 80.19\%  & \multicolumn{1}{c|}{\multirow{-2}{*}{{[}16,10,10,8{]}}} & \multicolumn{1}{c|}{\multirow{-2}{*}{DenseNet201}} & \multicolumn{1}{c|}{\multirow{-2}{*}{{\color[HTML]{32CB00} 0.26}}} & \multicolumn{1}{c|}{\multirow{-2}{*}{16.13\%}} & \multicolumn{1}{c|}{\multirow{-2}{*}{{\color[HTML]{32CB00} 0.35}}} & \multicolumn{1}{c|}{\multirow{-2}{*}{2.78\%}}  & \multicolumn{1}{c|}{\multirow{-2}{*}{81.93\%}} & \multirow{-2}{*}{0.10\%} \\ \hline
                                        & \multicolumn{1}{c|}{}                       & \multicolumn{1}{c|}{}                       &                           & \multicolumn{1}{c|}{{\color[HTML]{FE0000} 0.27}} & \multicolumn{1}{c|}{{\color[HTML]{FE0000} 0.43}} & 80.84\%  & \multicolumn{1}{c|}{{\color[HTML]{FE0000} 0.27}} & \multicolumn{1}{c|}{{\color[HTML]{FE0000} 0.49}} & 80.69\%  & \multicolumn{1}{c|}{}                                   & \multicolumn{1}{c|}{}                              & \multicolumn{1}{c|}{{\color[HTML]{32CB00} }}                       & \multicolumn{1}{c|}{}                          & \multicolumn{1}{c|}{{\color[HTML]{32CB00} }}                       & \multicolumn{1}{c|}{}                          & \multicolumn{1}{c|}{}                          &                          \\ \cline{5-10}
\multirow{-2}{*}{ResNet-18}             & \multicolumn{1}{c|}{\multirow{-2}{*}{0.26}} & \multicolumn{1}{c|}{\multirow{-2}{*}{0.43}} & \multirow{-2}{*}{81.28\%} & \multicolumn{1}{c|}{{\color[HTML]{FE0000} 0.38}} & \multicolumn{1}{c|}{{\color[HTML]{32CB00} 0.41}} & 81.97\%  & \multicolumn{1}{c|}{{\color[HTML]{FE0000} 0.35}} & \multicolumn{1}{c|}{{\color[HTML]{32CB00} 0.34}} & 80.00\%  & \multicolumn{1}{c|}{\multirow{-2}{*}{{[}16,10,16,8{]}}} & \multicolumn{1}{c|}{\multirow{-2}{*}{DenseNet121}} & \multicolumn{1}{c|}{\multirow{-2}{*}{{\color[HTML]{32CB00} 0.24}}} & \multicolumn{1}{c|}{\multirow{-2}{*}{7.69\%}}  & \multicolumn{1}{c|}{\multirow{-2}{*}{{\color[HTML]{32CB00} 0.39}}} & \multicolumn{1}{c|}{\multirow{-2}{*}{9.30\%}}  & \multicolumn{1}{c|}{\multirow{-2}{*}{82.10\%}} & \multirow{-2}{*}{0.82\%} \\ \hline
\end{tabular}
  \label{tab1}%
   \begin{tablenotes}
        \scriptsize
        \item  $\bullet$ {\color[HTML]{32CB00}Green} indicates successful optimization of the unfairness score. $\bullet$ {\color[HTML]{FE0000} Red} indicates failed optimization of the unfairness score.
      \end{tablenotes}
\end{table*}

\section{Experiments}

Muffin is evaluated on two datasets for diagnosing dermatological disease. Results show the superiority of
Muffin-Nets over the existing neural architectures since it can improve multi-dimensional fairness without compromising accuracy.


\noindent\textbf{4.1. Experimental Setup}

\textbf{\textit{A. Dataset:}} Two datasets  ISIC2019 \cite{ISIC2019} and  Fitzpatrick17K \cite{groh2021evaluating}
are built based on patients' images collected from the open-access datasets. The images from ISIC2019 are utilized for a classification task with 8 dermatology diseases and  Fitzpatrick17K is used for 9 classes of classification.

\textbf{\textit{B. Muffin settings:}} 
During the whole process, datasets were split into three sets: (1) training set
with 64\% images; (2) validation set with 16\% images; and (3) test set
with the rest 20\% images. The number of episodes for reinforcement
learning is set to 500.

\textbf{\textit{C. Competitors and training settings:}} 
To evaluate Muffin-Nets, we select a set of state-of-the-art neural networks for comparison, including (1) the manually designed ResNet \cite{targ2016resnet}, and (2) the AutoML identified MobileNetV3 \cite{Howard_2019_ICCV}, DenseNet \cite{huang2017densely} and ShuffleNet \cite{zhang2018shufflenet}.
For a fair comparison, all neural networks are trained from scratch with the same hyperparameters on a GPU cluster with 48 RTX 3080: (1) learning rate starts from 0.1 with a decay of 0.9 in 20 steps, (2) 64 for the batch size, and (3) 500 epochs for training.

\vspace{2pt}
\noindent\textbf{4.2 Comparison of Muffin with existing fairness Techniques for different models}

In Table \ref{tab1}, we compare Muffin-Nets against basedlines with 2 optimization methods (Data-based (D) and Loss-function based (L)). Considering the model size, we presented 4 different model types from the smallest one ShuffleNet\_V2\_X1\_0 to the biggest one ResNet-18.

We have several observations from Table \ref{tab1}: 
(1) Both existing optimization methods are not consistent for the models listed here. If we optimize one attribute, it will lead to an increase in unfairness for another attribute.
The data method can improve the accuracy with some architectures. But when comes to the loss function method, it declines the accuracy to a certain degree. 
(2) Compared with the above methods, Muffin can significantly improve the accuracy of small architectures. The parameters of ShuffleNet\_v2\_X1\_0 and
MobileNet\_V3\_Samll is 1261804 and 1526056 respectively, the overall accuracy is below 80\% even with the optimization methods, when it unites another model by Muffin framework, the accuracy improvement can reach 3.34\% and 5.58\%. As the big models like DenseNet121 and ResNet-18, Muffin still has a slight improvement of overall accuracy during the process. 
(3) Muffin can optimize two attributes at a time. It means we don't need to do the tradeoff between several fairness-related attributes. The improvement for fairness of age and site can reach 19.44\%, 2.22\%  for ShuffleNet\_v2\_X1\_0, 26.32\%, and 20.37\%  for MobileNet\_V3\_Samll.
When comes to DenseNet121 which meets the bottleneck of fairness for site, the enhancement can reach 2.78\% of site.
 Similarly, ResNet-18 which encounters the bottleneck of fairness for age, the improvement is 7.69\%.
(4) Compared with age, site is a much more difficult attribute to be optimized. With Muffin, the unfairness score of age can reach 0.22 but the lowest unfairness score of site is only 0.29. There are 6 subgroups for the age but 9 subgroups for the site, as the number of subgroups increases, it will be harder to make every subgroup at the same level.

\begin{figure}[t]
  \centering
  \includegraphics[width=3.4in]{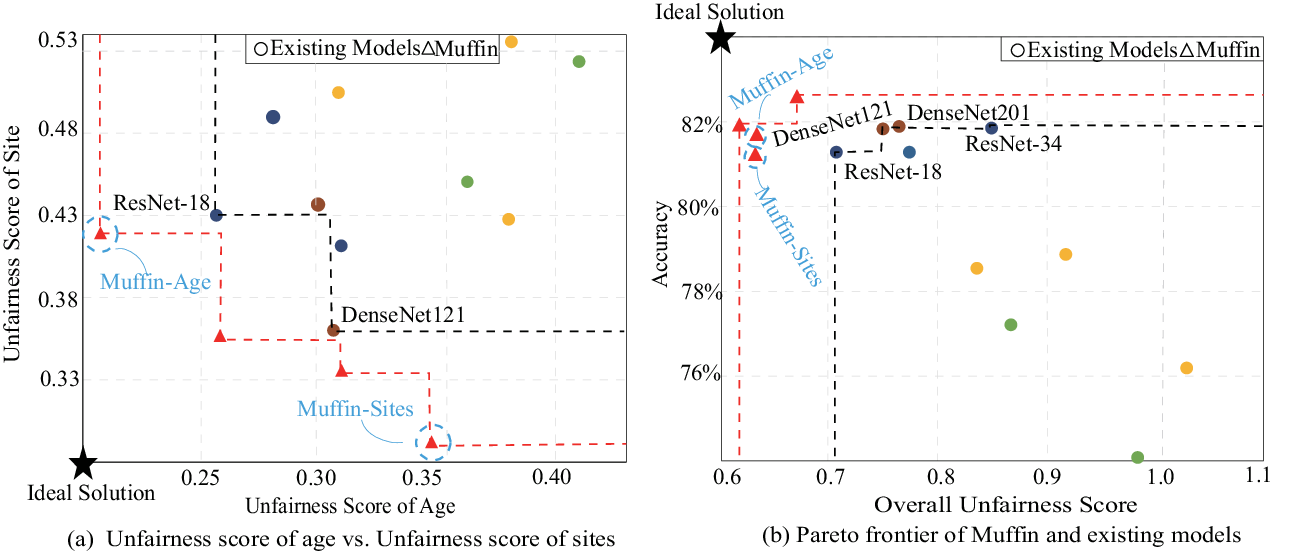}
  \caption{Comparison between the existing neural networks and
Muffin-Nets with the unfairness score and accuracy.}
    \vspace{-8pt}
  \label{fig:ex1}
\end{figure}

\vspace{2pt}
\noindent\textbf{4.3. Exploration by Muffin}

In the second set of experiments, we will demonstrate that Muffin-Nets can significantly push forward the Pareto frontiers among multiple attributes and accuracy for dataset.

\textbf{\textit{A. Unfairness score of sites vs. Unfairness score of age:}} In Figure \ref{fig:ex1}(a) the x-axis is the unfairness score of age and the y-axis is the unfairness score of site. The ideal solution is located in the left corner, denoted as a star. Each circled point corresponds to an existing network like Figure \ref{fig:mot1} and the red triangle is Muffin-Net which is generated by our framework.

In Figure \ref{fig:ex1}(a), The
red and black lines plot the Pareto frontiers of Muffin-Nets and
existing networks, respectively. We observe that
Muffin-Age dominates all the existing
neural networks in terms of unfairness score of age, it is only 0.2171; while Muffin-Sites located in the right corner achieves the highest fairness of the site.
These figures clearly show that Muffin can significantly push forward the Pareto frontiers considering the fairness of age and site.

\textbf{\textit{B. Accuracy vs. Unfairness.}} 
We further investigate the Pareto frontier between overall fairness and accuracy by integrating the unfairness score of age and site from Figure \ref{fig:ex1}(a).
Results in Figure \ref{fig:ex1}(b) consistently show that Muffin can also push forward the Pareto frontier considering accuracy and overall fairness.
More specifically, Muffin-Net is the only architecture whose accuracy can exceed 82\%. Muffin-Age and Muffin-Sites which lies in the Pareto frontier in Figure \ref{fig:ex1}(a) achieve accuracy of 81.71\% and 81.24\%, respectively.


\begin{figure}[t]
  \centering
  \includegraphics[width=3.4in]{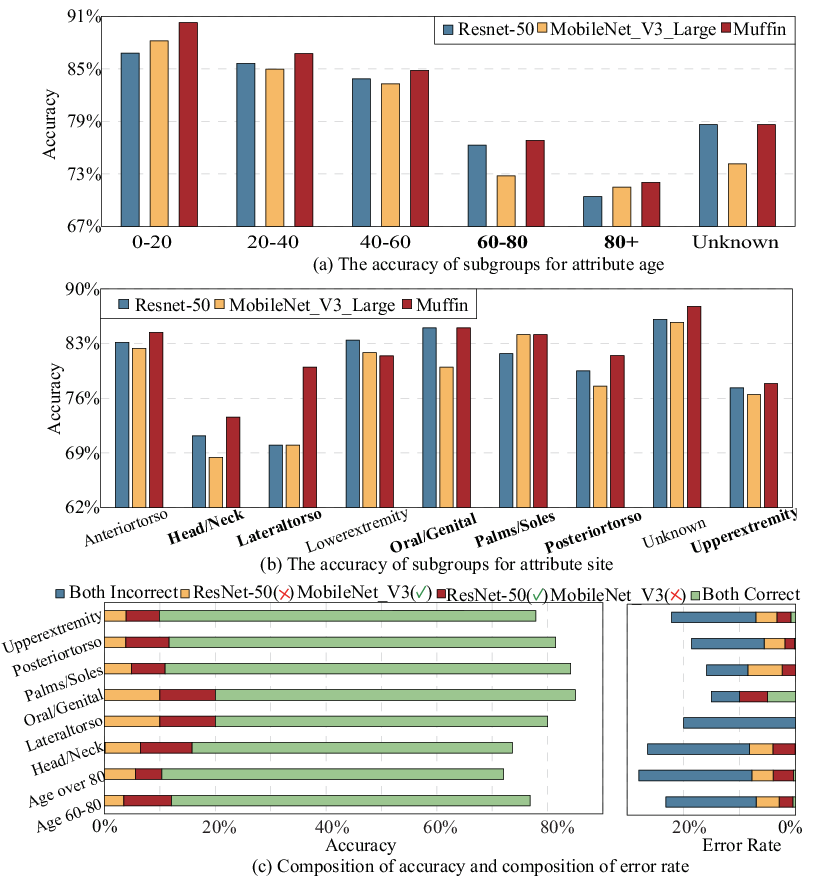}
  \caption{Detailed result of Muffin-Site on ISIC2019.}
    \vspace{-8pt}
  \label{fig:ex3}
\end{figure}

\vspace{2pt}
\noindent\textbf{4.4 Insights from Muffin-Site}

Figure \ref{fig:ex3} provided the accuracy of each sub-group, meanwhile, it gives out 
detailed composition of accuracy and error rate.

Muffin-site unites the ResNet-50 and MobileNet\_V3\_Large from the model pool. In the Figure \ref{fig:ex3}(a), 
when we look at
the privileged groups which are not indicated by bolded, Muffin has a slight rise compared with the highest accuracy from paired models. What's more, for the unprivileged groups which are bolded, the accuracy improvement of Muffin-site is more remarkable. As the accuracy gap between these two groups becoming smaller, it will lead to fairness improvement.

In Figure \ref{fig:ex3}(b), for all unprivileged groups considering the site, the accuracy is increasing especially the group lateraltorso.
The composition of accuracy and error rate for each unprivileged are presented in Figure \ref{fig:ex3}(c). The green bar which indicates that both models are correct is the main source of the accuracy for Muffin-Site. The yellow bar which represents MobileNet\_V3\_Large is correct and ResNet-50 is incorrect has the opposite meaning as the red bar.


To illustrate the situation even further, if we look at the site group lateraltorso, the 4th pillar from the bottom, Muffin-Site can fully leverage paired models. All images which can be judged correctly by either of the two models can be classified by Muffin-Site. There is no yellow or red part in the error rate composition. Another group named oral/Genital, Muffin-Site has done some incorrect classification thus there is green in the error composition. Nevertheless, all correct determination from ResNet-50 are kept by Muffin-Site and a portion of the right classification from MobileNet\_V3\_Large are also reserved by it. If we add these two parts, it will be far more than the green part which is moved to the error rate.
 As we hoped in \textbf{Observation 3}, Muffin can unite  both models perfectly. 



\vspace{2pt}
\noindent\textbf{4.5  Validation By Fitzpatrick17K}

\begin{figure}[t]
  \centering

  \includegraphics[width=3.4in]{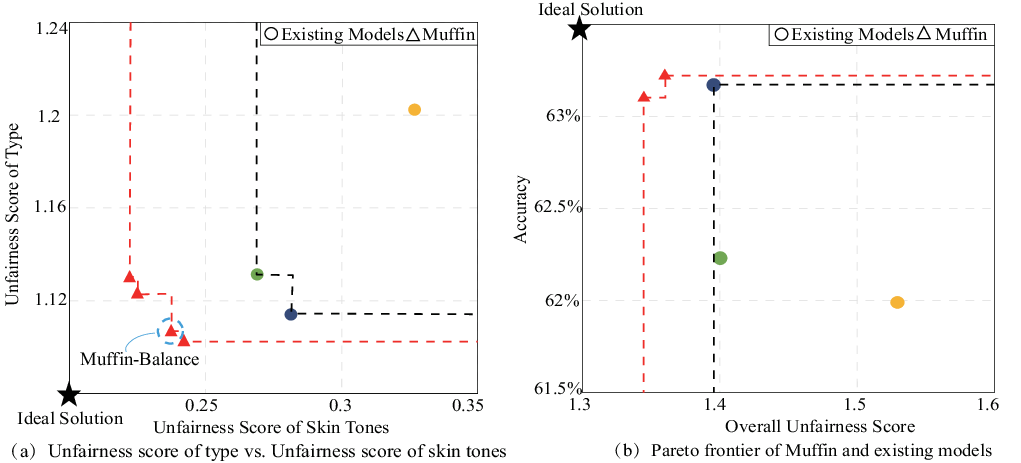}
 
  \caption{Validation of Muffin by Another Dataset Fitzpatrick17K.}
   \vspace{-22pt} 
  \label{fig:ex4}
\end{figure}

\begin{figure}[t]
  \centering

  \includegraphics[width=3.4in]{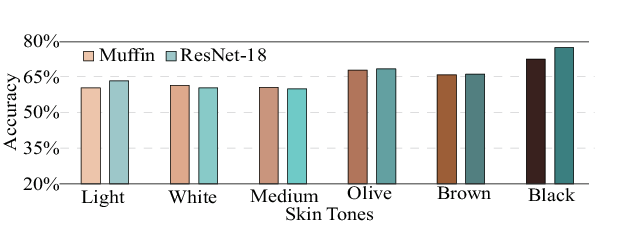}
 
  \caption{Detailed result of Muffin-Balance on Fitzpatrick17K.}
   \vspace{-10pt} 
  \label{fig:ex17}
\end{figure}

Based on dataset Fitzpatrick17K which also has the problem of multi-dimensional fairness, a model pool that has ResNet, ShuffleNet and MobileNet has been built. In Figure \ref{fig:ex4}(a), Muffin improves the unfairness score of skin tones and type significantly. Muffin-Balacen which lies in the Pareto frontiers considering the unfairness score of skin tones and type will be used for further illustration in Figure \ref{fig:ex17}.
In Figure \ref{fig:ex4}(b), Muffin also pushes forward Pareto frontiers
for overall unfairness and accuracy. This is good proof that Muffin is also useful for other datasets by leveraging models.

Fitzpatrick scale \cite{Fitzpatrick} has 6 types of skin color, from light to black. In Figure \ref{fig:ex17}, we have compared the accuracy for each sub-group of ResNet-18 which is located in the Pareto frontiers and Muffin. For the group white and medium, Muffin achieves accuracy gain but for black, Muffin has some loss of accuracy. In such a complementary way, the overall accuracy can be unaffected, while the model can become much fairer.




\vspace{2pt}
\noindent\textbf{4.6 Ablation Studies}

\begin{figure}[t]
  \centering
  \includegraphics[width=3.4in]{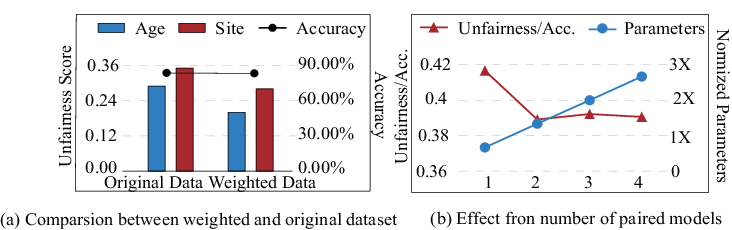}
  \caption{The importance of weighted data and number of paired models.}
    \vspace{-10pt}
  \label{fig:ex5}
\end{figure}

In Figure \ref{fig:ex5}(a), we compare the effect of training with the original dataset and the weighted dataset without unknown group for the same MLP structure. The paired model is optimized DesNet121 and original ResNet-18 and the MLP structure is {16,16,16,8}. It is obvious that with the weighted dataset, regardless age or site, the unfairness score will decline, and meanwhile, the overall accuracy can be kept.

In Figure \ref{fig:ex5}(b), 
for each number of paired models, two architectures that have the best performance will be selected to plot.
When it is equal to 1, ResNet-18 and DenseNet121 which locate in the Pareto frontier are selected. For Muffin, Muffin-Age, and Muffin-Sites are posted. 
If we expand the paired model into 3 or 4, we can still identify a series of models. It is obvious that even though we increase the number of paired models, the parameters will explode, but the reward is maintained at the same level.  This is a good illustration of the tradeoff between unfairness, accuracy, and parameters.

\section{Conclusion}

In this work, we have put multiple unfair attributes in a holistic optimization loop and proposed a framework, namely \textit{Muffin}. On top of it, a model fusing structure is built to unit the existing models and then  it will be trained with a proxy dataset in order to improve fairness. As such, Muffin can form a much better Pareto frontier on accuracy and multi-dimensional fairness compared with the existing models. As a whole, an automatic tool is developed to generate an AI system that can achieve multi-dimensional fairness.
Extensive experiments are carried out to evaluate Muffin, where it is consistent for all models' optimization and the maximum fairness improvement can reach 26.32\% for age and 20.37\% for the site. 


\section*{Acknowledgement}

We gratefully acknowledge the support of National Institutes of
Health (NIH) (Award No. 1R01EB033387-01).

\bibliographystyle{ieeetr}

{ 
\scriptsize
\bibliography{ref}}

\end{document}